\documentclass[abstract=true]{scrartcl}
\usepackage[margin=2.5cm]{geometry}

\usepackage{lineno,hyperref}
\usepackage[normal, bf]{caption}  
\usepackage{booktabs} 
\usepackage{natbib}
\usepackage{algorithmicx}
\usepackage{algorithm}
\usepackage{algpseudocode}
\usepackage{subfig}
\usepackage{xcolor}
\usepackage{graphicx}

\newcommand{\code}[1]{\texttt{#1}}

\begin{document}
	
\title{Instruction-Level Design of Local Optimisers using Push GP\footnote{Citation: M. Lones, Instruction-Level Design of Local Optimisers using Push GP, \textit{Genetic and Evolutionary Computation Conference Companion (GECCO'19)}, ACM, New York, 2019. ISBN 978-1-4503-6748-6/19/07. This is the author's own version. The published version is available at DOI: \href{https://doi.org/10.1145/3319619.3326806}{10.1145/3319619.3326806}.}}

\author{Michael A. Lones\footnote{School of Mathematical and Computer Sciences, Heriot-Watt University, Edinburgh, UK, \href{mailto:m.lones@hw.ac.uk}{m.lones@hw.ac.uk}.}}

\date{}
\maketitle

\begin{abstract}
This work uses genetic programming to explore the design space of local optimisation algorithms. Optimisers are expressed in the Push programming language, a stack-based language with a wide range of typed primitive instructions. The evolutionary framework provides the evolving optimisers with an outer loop and information about whether a solution has improved, but otherwise they are relatively unconstrained in how they explore optimisation landscapes. To test the utility of this approach, optimisers were evolved on four different types of continuous landscape, and the search behaviours of the evolved optimisers analysed. By making use of mathematical functions such as tangents and logarithms to explore different neighbourhoods, and also by learning features of the landscapes, it was observed that the evolved optimisers were often able to reach the optima using relatively short paths.
\end{abstract}

\section{Introduction}
There are many optimisation algorithms that can potentially be applied to a given problem, but theoretical understanding \citep{wolpert1997no, joyce2018review} tells us that a particular optimiser will not be effective over all problems. Despite some progress \citep{kerschke2018automated}, theory has not yet reached the stage where it can offer concrete guidance on which optimisers are suitable for particular problem types. This means that, in practice, it is often necessary to go through a process of trying out different optimisers to see which one is effective on a particular problem. Given that there are many optimisation algorithms available, and that each algorithm has many variants and potential hybrids, this process of selecting the optimal optimiser has the capacity to be very involved and time consuming. One way of addressing this is to use a machine learning algorithm to select or design an optimiser on the user's behalf \citep{rice1976algorithm, kerschke2019automated}. In particular, the hyperheuristics \citep{burke2013hyper,swan2018re} community has been exploring this idea for some time, typically by using evolutionary algorithms to select or generate the heuristics used by a particular metaheuristic framework, or generating new metaheuristic frameworks by combining existing heuristics.

This paper considers the problem of using genetic programming (GP) to design entire optimisers from scratch. In this respect, it is closely related to previous work on hyperheuristics, but with a focus on low-level design of novel search strategies, rather than specialising existing heuristic and metaheuristic algorithms. Another difference from most hyperheuristic approaches is that this work focuses on continuous, rather than discrete, optimisation. However, the aim of this work is not just to design better optimisers, but also to explore and understand the optimiser design space in a more systematic and context-sensitive manner. This is motivated by recent growth in the design of nature-inspired optimisation algorithms, which has seen the invention of new optimisers based on principles of animal foraging and other natural phenomena that are only tangentially related to optimisation \citep{sorensen2015metaheuristics,lones2019mitigating}. This work uses Push \citep{spector2001autoconstructive, spector2002genetic, spector2004push}, a language that was designed to address the need for both expressiveness and evolvability when optimising programs using evolutionary algorithms. Both of these properties are likely to be important when evolving complex behaviours such as optimisation from scratch. Push is also notable for having been designed for the related problem of auto-constructive evolution.

In this paper, the aim is to gain some understanding of the potential for using Push to design optimisation algorithms, rather than carrying out a rigorous experimental investigation. Consequently, the focus is on evolving local optimisers (which tend to be simpler than population-based approaches) on a small selection of problems that represent common types of solution landscapes, and then analysing the resulting optimisers. The paper is structured as follows: Section \ref{sec:related} discusses related work; Section \ref{sec:push} gives an introduction to Push; Section \ref{sec:methods} gives an overview of the methodology used in this paper; Section \ref{sec:results} presents results and analysis; Section \ref{sec:conclusions} concludes.

\section{Related Work} \label{sec:related}
In machine learning, it is well known that different algorithms perform well on different problems. In the field of optimisation, this notion has been formalised by various \textit{no free lunch} theorems \citep{wolpert1997no, joyce2018review} which prove that no optimiser is superior to any other when its performance is averaged over all possible problems. The task of an optimisation practitioner is therefore to find an algorithm that works well for a particular problem. \citet{rice1976algorithm}, who was the first to formalise this idea within the broader context of machine learning, referred to it as the \textit{algorithm selection problem}. One way to approach this problem is to treat it as a predictive modelling problem, i.e. training a model that can predict how well a particular algorithm will perform when it is provided with information about the problem domain. Such approaches are generally known as metalearning \citep{vilalta2002perspective}, or \textit{learning to learn}, and make use of metadata about previous experience using algorithms in order to train the predictive model. For example, this approach has been widely researched in the domain of data mining, where the aim is to use it to select an appropriate classification or regression algorithm for a given data set \citep{vilalta2004using}.

Within the optimisation community, previous work on algorithm selection has focused on fitting specific optimisers to specific problems. That is, rather than choosing between different algorithms, the aim is to specialise a given optimiser in some way so that it is better adapted to a particular problem domain, or even a specific problem instance. In the simplest case, this can amount to parameter tuning (also known as hyperparameter optimisation). However, in the more general case, and particularly where significant behavioural changes are made to the optimiser, this approach has come to be known as \textit{hyperheuristics} \citep{burke2013hyper, swan2018re} (\citep{pappa2014contrasting} provides a good review of the overlap between hyperheuristics and metalearning). Early work in this area focused on \textit{selective} hyperheuristics, where the metalearner is used to choose between existing heuristic components of a metaheuristic framework, sometimes in a dynamical fashion. A more recent approach, known as \textit{generative} hyperheuristics, involves using GP to make changes to a metaheuristic algorithm's underlying code, with the aim of creating new heuristics that are fitted to a particular problem's search space. Examples are the design of new mutation \citep{woodward2012automatic}, crossover \citep{goldman2011self} and selection \citep{richter2018automated} operators for evolutionary algorithms.

There are also examples of using GP to design the overall structure of metaheuristic optimisers, typically by combining components of existing metaheuristic frameworks in novel ways. In some cases this approach has been used to design new kinds of optimiser \citep{martin2013evolving, ryser2016iterative}, and in other cases new versions of existing algorithms, e.g. evolution strategies \citep{van2016evolving} and Nelder-Mead \citep{fajfar2017evolving}. Various forms of GP have been used for this, including tree-based GP \citep{martin2013evolving, fajfar2017evolving}, Cartesian GP \citep{ryser2016iterative}, linear GP \citep{oltean2005evolving}, and grammatical evolution \citep{lourencco2012evolving}.

More recently, the deep learning community has also begun to address the problem of automatically designing optimisers, particularly those used for training neural networks \citep{li2017learning, wichrowska2017learned}. Broadly speaking, the idea is similar to the one explored in this paper: training computational systems that, based on inputs describing the state of the search process, generate outputs (and internal state) that determine subsequent steps in the process. However, a key difference is that the computational representation used in this work is closer to the one used by human programmers, and therefore more amenable to human understanding.

\section{Push} \label{sec:push}

\begin{table*}[tb!]
	\centering
	\small
	\caption{Vector stack instructions}
	\label{tab:vector}
	\begin{tabular}{llll}
		\toprule
		Instruction&Pop from&Push to&Description\\
		\midrule
		\code{vector.+}&vector, vector&vector&Add two vectors\\
		\code{vector.-}&vector, vector&vector&Subtract two vectors\\
		\code{vector.*}&vector, vector&vector&Multiply two vectors\\
		\code{vector./}&vector, vector&vector&Divide two vectors\\
		\code{vector.scale}&vector, float&vector&Scale vector by scalar\\
		\code{vector.dprod}&vector, vector&float&Dot product of two vectors\\
		\code{vector.mag}&vector&float&Magnitude of vector\\
		\code{vector.dim+}&vector, float, int&vector&Add float to specified component\\
		\code{vector.dim*}&vector, float, int&vector&Multiple specified component by float\\
		\code{vector.apply}&vector, code&vector&Apply code to each component in turn\\
		\code{vector.zip}&vector, vector, code&vector&Apply code to each pair of components in turn\\
		\code{vector.rand}&&vector&Generate a random vector of floats\\
		\code{vector.urand}&&vector&Generate a random unit vector\\
		\code{vector.wrand}&&vector&Generate a random vector within the bounds\\
		\code{vector.between}&vector, vector, float&vector&Gets point between two vectors, with the offset\\
		&&&given by a scalar value\\
		\bottomrule
	\end{tabular}
\end{table*}

Push \citep{spector2001autoconstructive, spector2002genetic, spector2004push} is a  stack-based typed language designed for use within GP. A Push program is a list of instructions, each of which operates upon a specified type stack. There are stacks for primitive data types (booleans, floats, integers) and each of these has both special-purpose instructions (e.g. arithmetic instructions for the integer and float stacks, logic operators for the boolean stack) and general-purpose stack operators (push, pop, swap, duplicate, rot etc.). There is a code stack, which allows code to be stored and later executed, and a Push program can modify its own code during execution using instructions that operate on its execution stack. Finally, there is an input stack, which remains fixed during execution. This provides a way of passing non-volatile constants to a Push program; when popped from the input stack, corresponding values get pushed to the appropriate type stack.

One of the primary benefits of using a stack-based approach over more traditional tree-based representations is that the actions of instructions are partially decoupled from their location within the genome. In tree-based GP, the effect of an instruction upon execution is determined by its location within a parse tree. By comparison, in a Push program there is a degree of decoupling between the location of an instruction or argument within the representation and its actual role within the program's execution. This occurs because arguments and results can be stored within stacks and used later; or not used at all. The consequence of this is that programs are less likely to be disrupted by mutation operators. Because the stack system decouples arguments from the functions which use them, it is also not possible to apply incorrectly-typed arguments to a function; therefore, there is no need to use special syntax-preserving recombination operators.

This work uses a modified version of the \code{Psh}\footnote{\url{http://spiderland.org/Psh/}} interpreter, which is an existing Java implementation of the Push language. The most significant change made to the code is the addition of a vector stack. This stores fixed-length floating point vectors, which can be used to represent search points. A range of special-purpose instructions have been defined for this stack; these are shown in Table \ref{tab:vector}. The \code{vector.apply} and \code{vector.zip} instructions allow code to be applied to each component (or each pair of components in the case of zip) using a functional programming style. The \code{vector.between} instruction returns a point on a line between two vectors. For this instruction, the distance along the line is determined by a value popped from the float stack; if this value is between 0 and 1, then the point is a corresponding distance between the two vectors; if less than 0 or greater than 1, then the line is extended beyond the first or second vector, respectively.

\section{Evolving Local Optimisers} \label{sec:methods}

\begin{algorithm}[tb!]
	\caption{Evaluating a Push optimiser}
	\label{alg::evaluation}
	\begin{algorithmic}[1]
		\For{$i \gets 1,\mathit{repeats}$}
		\State $p_i \gets$ random point within search bounds
		\State $v_i \gets \textsc{value}(p_i)$,\hspace{2mm}$\mathit{best}_i \gets$ infinity,\hspace{2mm}$v_i^{m-1} \gets$ infinity
		\State clear all stacks
		\State \textsc{push}(bounds, \code{input})
		\State \textsc{push}($p_i$, \code{vector} and \code{input})
		\State \textsc{push}($f$, \code{float} and \code{input})
		\State \textsc{push}(\code{true}, \code{boolean})
		\For{$m \gets 1,\mathit{moves}$}
		\State \textsc{push}($m$, \code{integer})
		\State \textbf{$<$ execute evolved Push expression $>$}
		\State $p_i \gets$ \textsc{pop}(\code{vector})
		\State $v_i^{m-1} \gets v_i$,\hspace{2mm}$v_i \gets \textsc{value}(p_i)$
		\State \algorithmicif \hspace{0.5mm} $v_i < \mathit{best}_i$ \hspace{0.2mm} \algorithmicthen \hspace{0.5mm} $\mathit{best}_i \gets v_i$
		\State clear input stack
		\State \textsc{push}(bounds, \code{input})
		\State \textsc{push}($v_i$, \code{float} and \code{input})
		\If{$p_i$ is within bounds}
			\State \textsc{push}($v < v^{m-1}$, \code{boolean})
			\State \textsc{push}($p_i$, \code{input})
			\State \textsc{push}($v_i$, \code{float} and \code{input})
			\State \textsc{push}($\mathit{best_i}$, \code{vector})
		\Else
			\State \textsc{push}(\code{false}, \code{boolean})
			\State \textsc{push}(infinity, \code{float})
		\EndIf
		\EndFor
		\EndFor
		\State \textit{fitness} $\gets$ \textsc{mean}($\mathit{best}_1,\ldots,\mathit{best}_{\mathit{repeats}}$)

	\end{algorithmic}
\end{algorithm}

A local optimiser carries out a series of moves within a search space of candidate solutions, with the aim of finding the candidate solution that minimises (or maximises) a particular objective function. Well-known local optimisation algorithms include hillclimbing, simulated annealing and tabu search. All of these typically begin the optimisation process with a random solution. At each iteration, they then make (typically small) changes to their current solution in order to sample a new solution. This new solution is then accepted according to some algorithm-specific criteria. For example, a hill-climber always accepts a new solution that has a better objective value than the current solution, simulated annealing accepts an improved solution in a probabilistic manner, and tabu search also takes into account prior search experience. If a new solution is accepted, a move takes place and the new solution becomes the current solution.

In this work, evolved Push expressions are used to generate moves. The aim is to carry out a broad exploration of local search strategies, so no constraints are placed on the manner in which moves are generated and/or accepted. However, some assistance is provided. First of all, the Push expression is called in a loop, meaning that it only needs to generate a single move each time it is called and consequently does not need to evolve its own outer loop. Second, before the Push expression is called, the objective value of the search point on the top of the \code{vector} stack is calculated and is pushed to the \code{float} stack. If this value is better than the previous time a search point was evaluated, \code{true} is pushed to the \code{boolean} stack; otherwise \code{false} is pushed. This means that the Push expression can easily check whether the last move was improving. Finally, certain information is always available on the input stack: namely, the current search point, its objective value, and the bounds of the search space. See Algorithm \ref{alg::evaluation} for more information.

This approach is evaluated on four functions taken from the CEC 2005 real-valued parameter optimisation benchmarks \citep{suganthan2005problem}. These are all minimisation problems, meaning that the aim is to find the input vector (i.e. search point) that generates the lowest value when passed as an argument to the function. The Sphere function ($F_1$) has a single optimum sitting in a bowl that curves upwards in all directions from the optimum. Schwefel's function ($F_2$) has a single optimum sitting in a valley between two peaks; it is harder than F1 because it is non-separable, so the dimensions can not be treated independently. Rosenbrock's function ($F_6$) is multi-modal and has a very narrow valley leading from the local optimum to the global optimum. Rastrigin's function ($F_9$) is also multi-modal, but has a large number of regularly spaced local optima.

Each time the fitness of a Push optimiser is measured, it is evaluated ten times from ten randomly-generated starting points located within the search space bounds specified in \citep{suganthan2005problem}. Its fitness is then the mean of the lowest objective values (i.e. best solutions) found in each of these optimisation runs. The use of multiple runs from random points prevents the optimiser from over-learning a particular part of the search space, and also gives a more robust measure of fitness. However, evaluation is very expensive. To make the experiments tractable, results are only collected for the 10-dimensional versions of the benchmark problems with a limit of 1000 moves per optimisation run. Table \ref{table:settings} shows the \code{Psh} parameter settings used in this study.

\begin{table}[!tb]
	\centering
	\caption{Psh parameter settings}
	\begin{tabular}{p{15.5cm}}
		\toprule
		Population size  = 200\\
		Maximum generations = 50\\
		Tournament size = 5\\
		Program size limit = 100\\
		Execution limit = up to 100 instruction executions per move\\
		Instruction set: \\
		\code{\textbf{boolean/float/integer/vector.} dup flush pop rand rot shove stackdepth swap yank yankdup}\\
		\code{\textbf{boolean.} = and fromfloat frominteger not or xor}\\
		\code{\textbf{exec.} = do*count do*range do*times if iflt noop}\\ 
		\code{\textbf{float.} \% * + - / < = > abs cos erc exp fromboolean frominteger ln log max min neg pow sin tan}\\
		\code{\textbf{input.} inall inallrev index}\\
		\code{\textbf{integer.} \% * + - / < = > abs erc fromboolean fromfloat ln log max min neg pow}\\
		\code{\textbf{vector.} * / + - apply between dim+ dim* dprod mag pop scale urand wrand zip}\\
		\code{false true}\\
		\bottomrule
	\end{tabular}
	\label{table:settings}
\end{table}

\section{Results} \label{sec:results}

\begin{figure}[tb!]
	\centering
	\includegraphics[width=10cm]{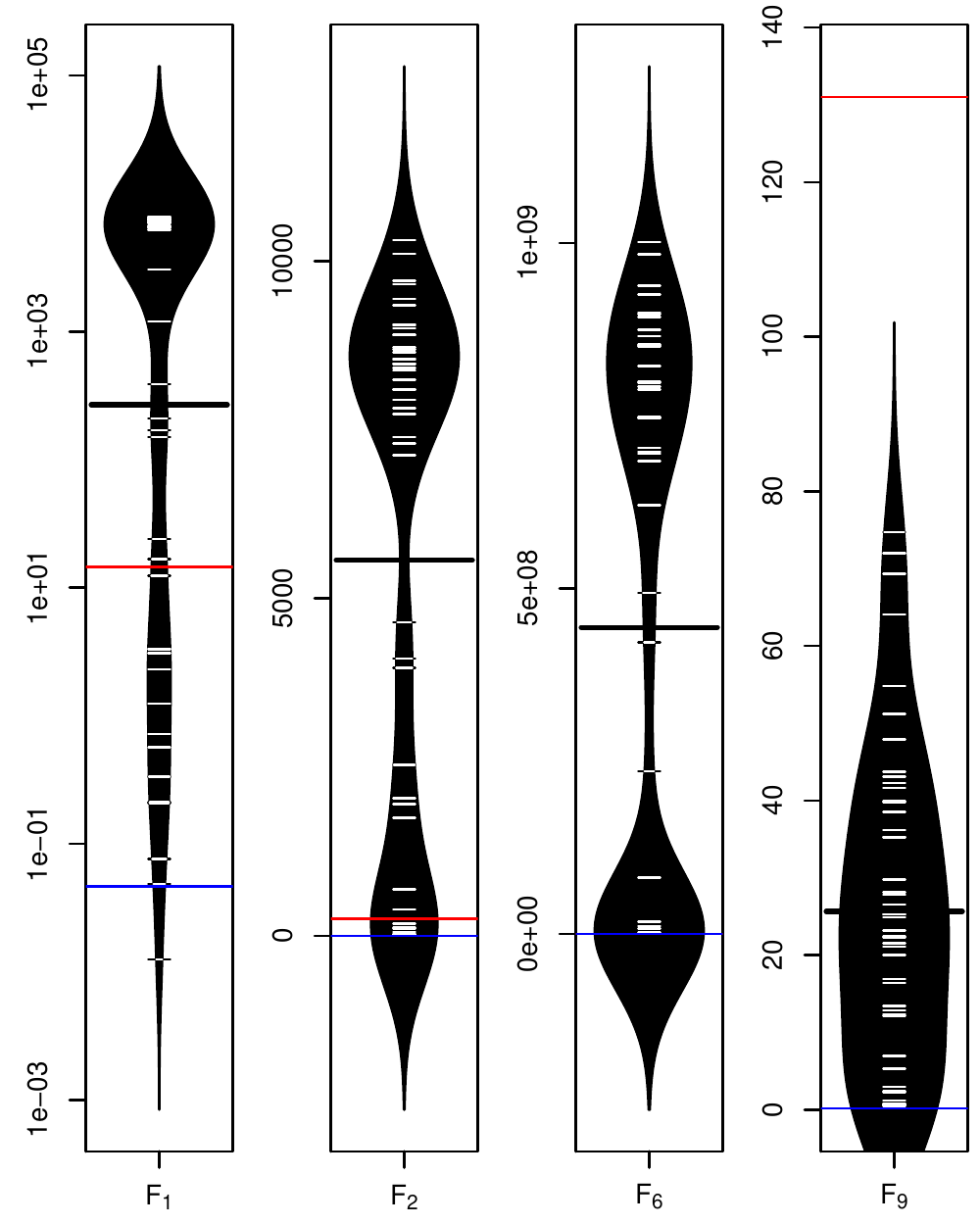}
	\caption{Distributions of mean errors for the best solutions from 50 Push GP runs for each benchmark. The best solution from each was re-evaluated from 25 random starting points and mean errors for these are shown as horizontal blue lines. Red lines show results reported in \citep{auger2005performance} (note, for $F_6$, the two lines are overlapping).}
	\label{fig:results}
\end{figure}

Figure \ref{fig:results} plots the fitness distributions over 50 runs of Push GP for each benchmark function. This shows that in each case multiple optimisers were found that could get close to the optimum objective value. In general, the distributions are wide and mostly bimodal, indicating that a significant proportion of runs were not successful at finding functional optimisers. The easiest problem was the unimodal Sphere function ($F_1$), with 16 evolved optimisers able to get very close to the optimum. The highly-multimodal Rastrigin's function ($F_9$) was also relatively easy, with 8 near-optimal solutions. Schwefel's function ($F_2$) proved much more challenging, with only 2 evolved optimisers getting close. Rosenbrock's function ($F_6$) was the most challenging, with no optimisers making it all the way to the optimum within the limit of 1000 moves.

To give an idea of how the error rates shown in Figure \ref{fig:results} compare to those achieved by an existing optimiser, error rates (taken from \citep{auger2005performance}, with the same dimensionality and maximum number of moves) are also shown for a version of CMA-ES. It can be seen that the evolved optimisers compare favourably on these problems, getting closer to the optima than CMA-ES in all cases (i.e. blue lines are below red lines in Figure \ref{fig:results}). However, this is not unexpected, since a specialist optimiser trained on a particular function landscape would be expected to do better than a general-purpose optimiser. It should also be borne in mind that a large evaluation budget has been consumed during the Push GP runs, so a direct comparison makes only limited sense.

\begin{table*}[htb!]
	\centering
	\caption{Examples of evolved Push optimisers with low error rates}
	\label{tab:optimisers}
	\begin{tabular}{rp{13.2cm}}
		\toprule
		Evolved on&Example optimiser\\
		\midrule
		$F_1$&\code{(input.index float.min 0.68 float.+ float.tan vector.pop vector.wrand vector.- 0.75)}\\
		$F_2$&\code{(vector.yank vector.pop vector.yank float.ln vector.wrand vector.- 0.89 vector.wrand vector.-)}\\
		$F_6$&\code{(float.tan vector.wrand vector.yank vector.pop vector.- 0.61 vector.wrand vector.-)}\\
		$F_9$&\code{(vector./ float.* vector./ float.sin vector.dim+ vector.swap float.rand)}\\
		$F_{1,2,6,9}$&\code{(float.tan vector.wrand) (code.noop (((code.noop)) (((integer.fromboolean code.noop))) input.index ((vector.+ () integer.\% ((exec.do*count) ((vector.shove) )))) vector.wrand  ((vector.wrand))))}\\
		\bottomrule
	\end{tabular}
\end{table*}

\subsection{Analysis of Evolved Optimisers}

These results strongly indicate that Push GP can be used to express and evolve useful optimisers. However, another aim of this study is to look at whether evolved optimisers can provide useful insights into ways of doing optimisation. To address this, Table \ref{tab:optimisers} gives an example of an optimiser evolved for each of the benchmark functions. In the cases of $F_1, F_2$ and $F_6$, these are the optimisers that produced the lowest errors in training. In the case of $F_9$, the optimiser with the lowest error proved too challenging to analyse (it comprised 74 instructions and 2 inner loops) so the example given is for the third best optimiser, which used considerably fewer instructions with only a slightly higher error rate.

The $F_1, F_2$ and $F_9$ optimisers shown in Table \ref{tab:optimisers} all make moves through the search space by adding a random vector to the previous best point. Given the relatively low move limit, this seems to be an effective way of exploring moves in multiple dimensions at once, and was generally preferred to making moves in individual dimensions in the best-performing optimisers. It is notable that both the $F_1$ and $F_6$ optimisers use the tangent function to determine the size of components within this random vector, and hence the neighbourhood explored in a single move. In both cases, the tangent function is applied to the current objective value, meaning that the neighbourhood size varies periodically as search progresses. The shape of the tangent function entails that move sizes are widely distributed, yet most of the time they are relatively small. This seems to offer a good trade-off between intensification and diversification during search.

\begin{figure*}[tb!]
	\hspace{3mm}
	\subfloat[Optimising $F_1$ (\href{https://youtu.be/fgd8HDaw9UY}{movie})] {
		\hspace{-4mm}
		\includegraphics[trim={0 1cm 0 2cm}, width=225pt]{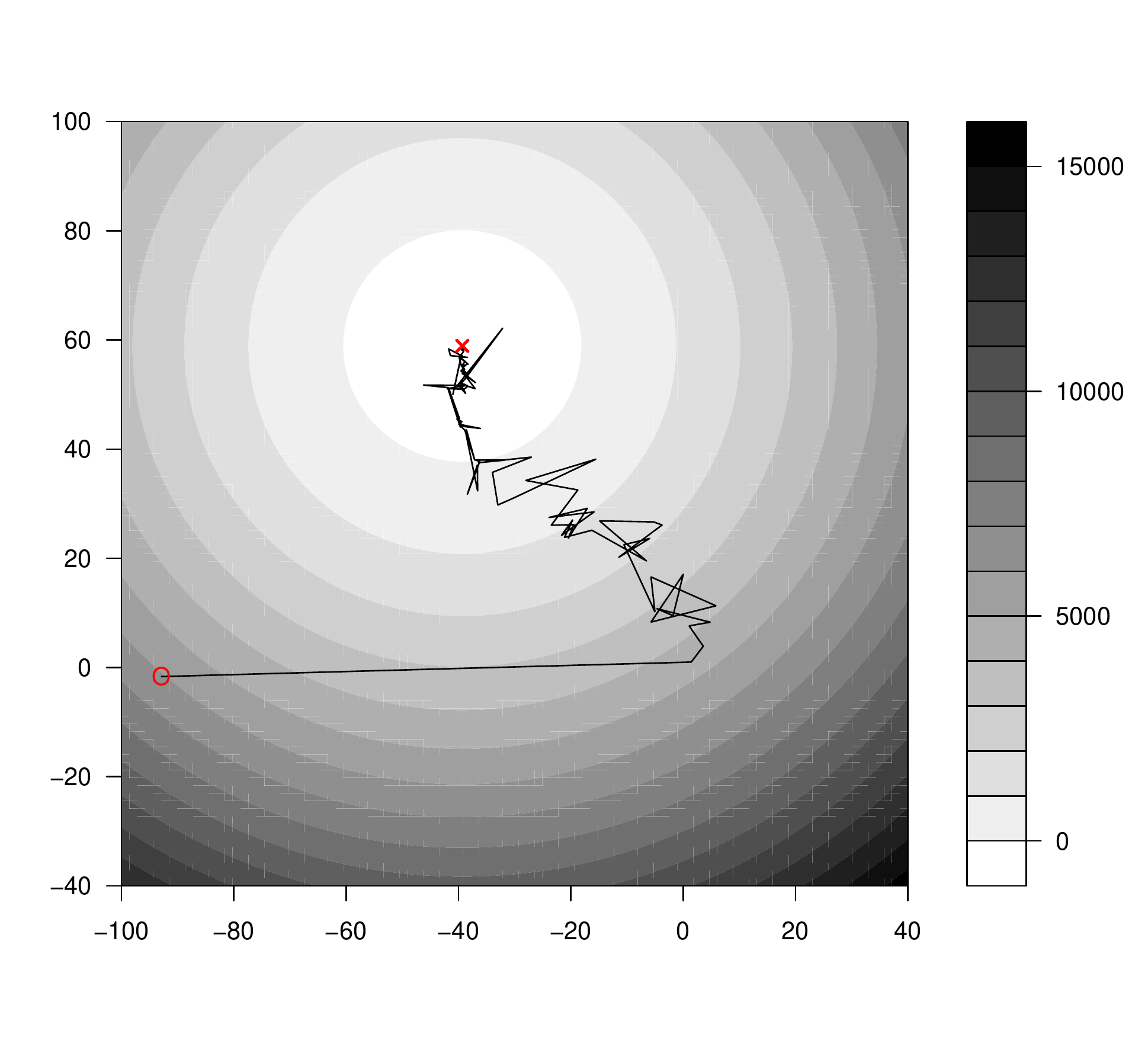}
		\label{fig:trajectories:f1}
	}
	\subfloat[Optimising $F_2$ (\href{https://youtu.be/fZ9hFAFeOgM}{movie})] {
		\includegraphics[trim={0 1cm 0 2cm}, width=225pt]{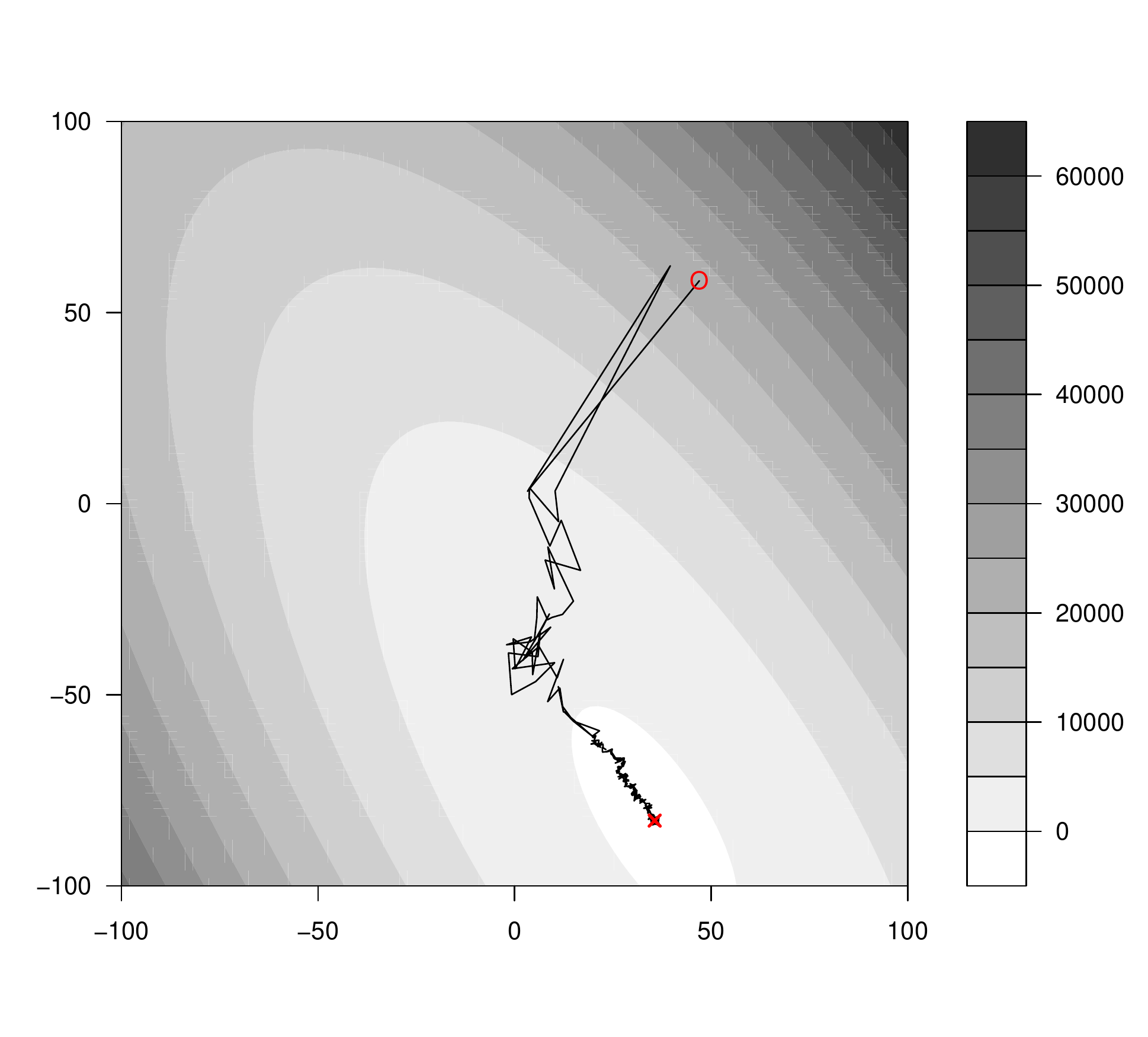}
		\label{fig:trajectories:f2}
	}\\
	\subfloat[Optimising $F_1$ using $F_2$ optimiser  (\href{https://youtu.be/Qek3HHYDpUc}{movie})] {
		\includegraphics[trim={0 1cm 0 2cm}, width=225pt]{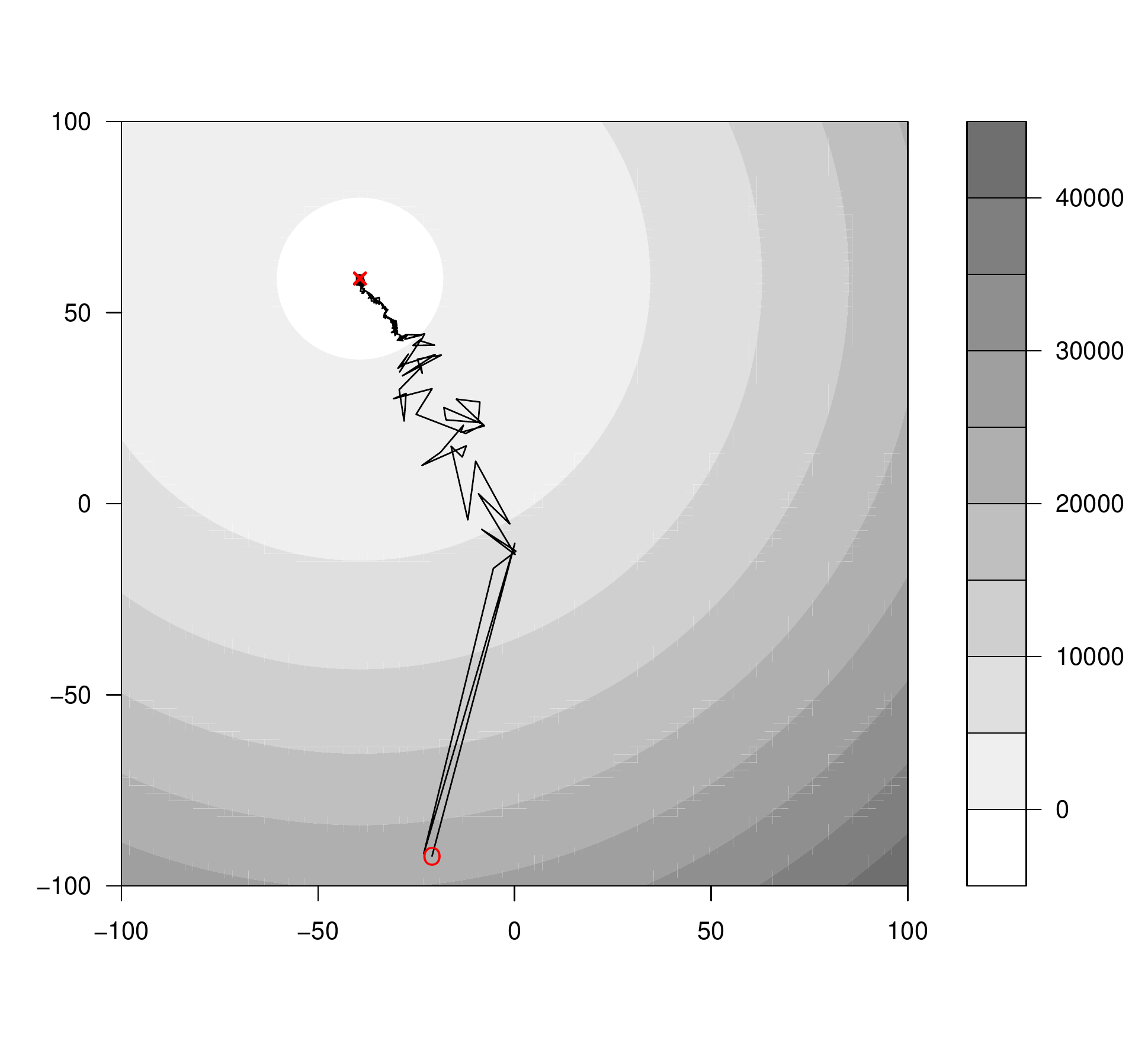}
		\label{fig:trajectories:f1f2}
	}
	\subfloat[Optimising $F_9$ (\href{https://youtu.be/gUinKW8iDG8}{movie})] {
		\includegraphics[trim={0 1cm 0 2cm}, width=225pt]{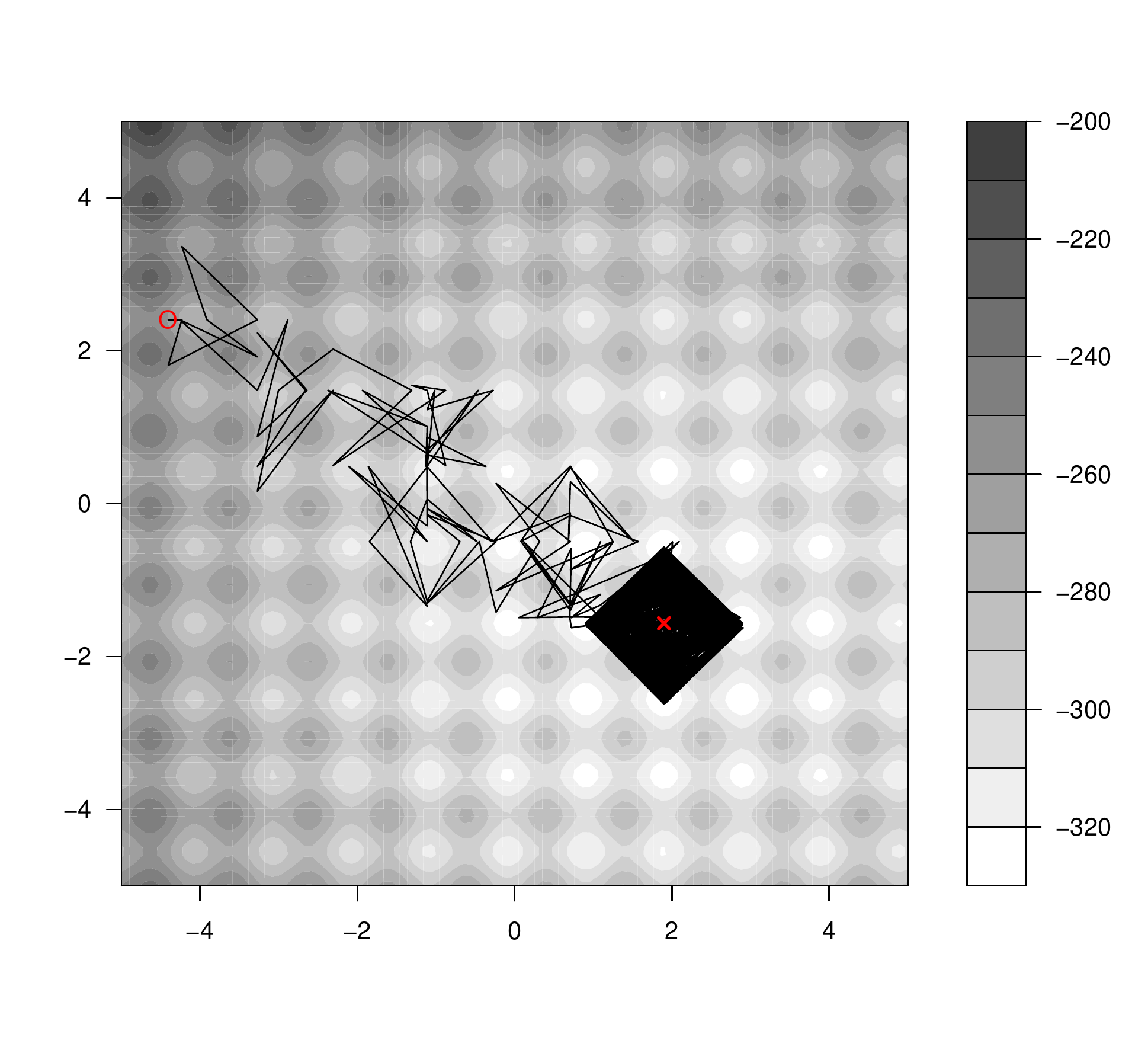}
		\label{fig:trajectories:f9}
	}
	\caption{Examples of search trajectories explored by evolved optimisers for 2-dimensional versions of the benchmark functions, with objective values shown as contours. Note that the optimisers were evolved on 10-dimensional versions of the functions. The initial search point is shown as a red circle and the best search point found is indicated by a red cross.}
	\label{fig:trajectories}
\end{figure*}

The $F_1$ optimiser appears to be particularly well-adapted to its search space. The first thing it does is  sample search points near the origin for several steps. If any of these is better than the initial random solution, it relocates search to this central region of the search space. This can be seen in Figure \ref{fig:trajectories:f1} after the first move. Given the curvature of the $F_1$ optimisation landscape, this will be the case whenever the initial solution is to the right of or below the origin, and hence this serves as an effective way of bootstrapping search. After this, and before the objective value reaches a threshold of 0.75, components of the random vector are sampled uniformly within the range $[-7.06,7.06]$, resulting in the large moves around the search space visible in Figure \ref{fig:trajectories:f1} from about $(0,0)$ to $(-40,40)$. Once it enters a higher value region, it then uses the tangent function to determine move sizes, as described above. This can be seen in Figure \ref{fig:trajectories:f1} after about $(-40,40)$, where the trajectory moves quite smoothly towards the optimum, with occasional large deviations when the tangent function grows large. Furthermore, the tangent function is aligned (by adding 0.68 to the objective value) so that moves become very small as it approaches the optimum objective value. Finally, if search goes out of bounds at any point, the move sizes become small (0.15) until it moves back within  bounds.

\begin{table}[tb!]
	\centering
	\caption{Generalisation to other functions}
	\label{table:generality}
	\begin{tabular}{rllllllll}
		\toprule
		Evolved on&$F_1$&$F_2$&$F_6$&$F_9$\\
		\midrule
		$F_1$&{\color{darkgray}4.67e-2}&2.29e+1&4.29e+4&5.18e+1\\
		$F_2$&7.57e-1&{\color{darkgray}4.75e+0}&1.33e+6&1.30e+2\\
		$F_6$&1.69e+0&9.83e+2&{\color{darkgray}1.65e+3}&5.94e+1\\
		$F_9$&1.71e+4&1.42e+4&4.61e+9&{\color{darkgray}1.82e-1}\\
		$F_{1,2,6,9}$&1.71e-1&1.27e+3&6.12e+3&5.38e+1\\
		\bottomrule
	\end{tabular}
\end{table}

\begin{table}[tb!]
	\centering
	\caption{Generalisation to other dimensionalities}
	\label{table:dimensionality}
	\begin{tabular}{rllllllll}
		\toprule
		Evolved on&2D&{\color{darkgray}10D}&15D&20D\\
		\midrule
		$F_1$&0.00e+0&{\color{darkgray}4.67e-2}&7.24e-1&6.32e+0\\
		$F_2$&8.51e-4&{\color{darkgray}4.74e+0}&7.34e+2&3.85e+3\\
		$F_6$&7.83e+1&{\color{darkgray}1.65e+3}&1.67e+4&9.87e+5\\
		$F_9$&4.12e-1&{\color{darkgray}1.82e-1}&1.23e+2&2.14e+2\\
		\bottomrule
	\end{tabular}
\end{table}

The $F_1$ optimiser uses a range of search heuristics. However, some of these are quite specific to the $F_1$ landscape, so it is interesting to consider whether it generalises to other problems. As an indication of this, Table \ref{table:generality} shows the error rates when the optimisers are reevaluated on the other three benchmark functions. It can be seen that reasonable generality is achieved on $F_2$, though this is also unimodal and has the same optimum objective value. It also generalises somewhat to $F_9$, with the large moves enabling it to move between local optima basins; however, the different scale of the search space means that it can not easily converge on the optima. No generality was observed when it was reevaluated on $F_6$, despite the fact that the $F_6$ optimiser in Table \ref{tab:optimisers} also uses the tangent function to determine move sizes. The $F_6$ optimiser, by comparison, does generalise well to $F_1$, but does not perform well on the other two functions.

The $F_2$ optimiser in Table \ref{tab:optimisers} uses the logarithm function rather than the tangent function to determine the neighbourhood size. This is adaptive in a more straightforward manner, with move sizes becoming smaller as search progresses towards the optimum. Two random vectors are added to the current best; the first is determined by the logarithmic function, the second with fixed limits of $[-0.89,0.89]$. Once the fitness drops below 0, only the latter random vector is added, and hence the neighbourhood size has a fixed bound. These behaviours can be seen in Figure \ref{fig:trajectories:f2}; the initial moves are very large, and become progressively smaller as the trajectory approaches the white area (where objective values are less than 0); after this, the move sizes become relatively small. When the trajectory goes out of bounds (not shown here) the magnitude of the random vector becomes very large, effectively causing a restart. According to Table \ref{table:generality}, this optimiser has good generality when re-evaluated on $F_1$; an example trajectory is depicted in Figure \ref{fig:trajectories:f1f2}. However, it performs poorly on the multimodal landscapes.

The $F_1, F_2$ and $F_6$ optimisers have broad similarities, particularly the way in which they adapt their move sizes as the objective value changes. Such strategies are likely to be effective in smooth landscapes with few optima, and variants of this idea are also seen in conventional optimisers. The $F_9$ optimiser, which must navigate a regular landscape with many local optima embedded within a broader curving gradient, behaves in a very different way. This can be seen in the trajectory shown in Figure \ref{fig:trajectories:f9}. The Push expression, although short, is difficult to understand behaviourally because the two vector divisions take into account values calculated in the previous moves. However, the trajectory shows that this iterative division, in concert with the sine function, causes fairly large movements which are broadly adapted to the distance between optima basins, allowing it to hop between neighbouring basins. After a while, it becomes attracted to, and carries out search within, a diagonal-shaped region comprising parts of five attractor basins. Even though their implementations vary markedly, all the $F_9$ optimisers with low error rates display broadly similar trajectories within the search space, and perform much better than CMA-ES. However, as Table \ref{table:generality} shows, they generalise very poorly to the other benchmark functions, suggesting that this search behaviour is quite specific to search spaces with regularly-spaced optima.

\subsection{Increasing Optimiser Generality}

\begin{figure*}[tb!]
	\subfloat[Optimising $F_1$, trained on $F_{1,2,6,9}$  (\href{https://youtu.be/Aevtz--alsI}{movie})] {
		\hspace{-4mm}
		\includegraphics[trim={0 1cm 0 2cm}, width=225pt]{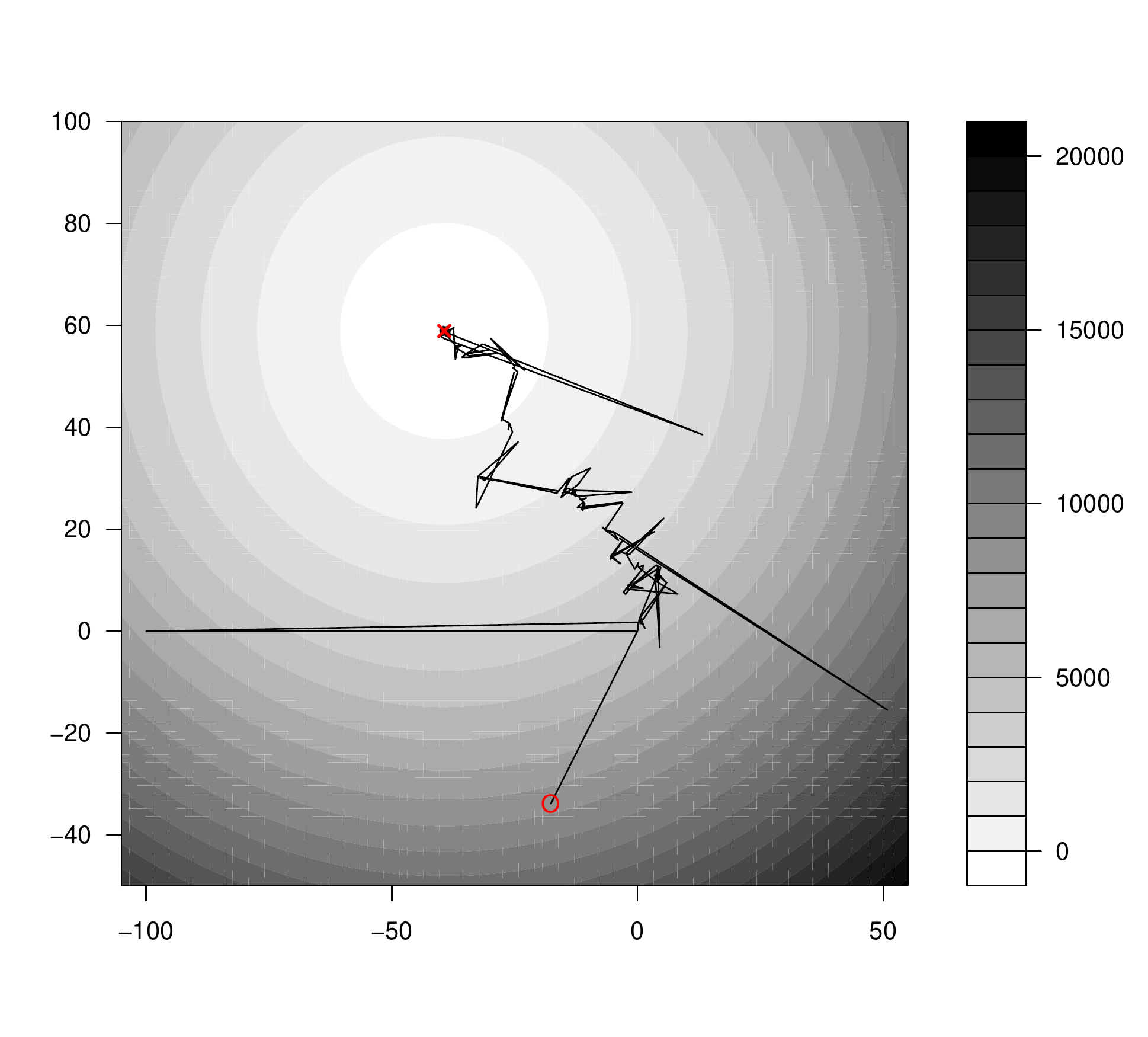}
		\label{fig:trajectories:f1269:f1}
	}
	\subfloat[Optimising $F_9$, trained on $F_{1,2,6,9}$  (\href{https://youtu.be/prmT8uLOqyE}{movie})] {
		\includegraphics[trim={0 1cm 0 2cm}, width=225pt]{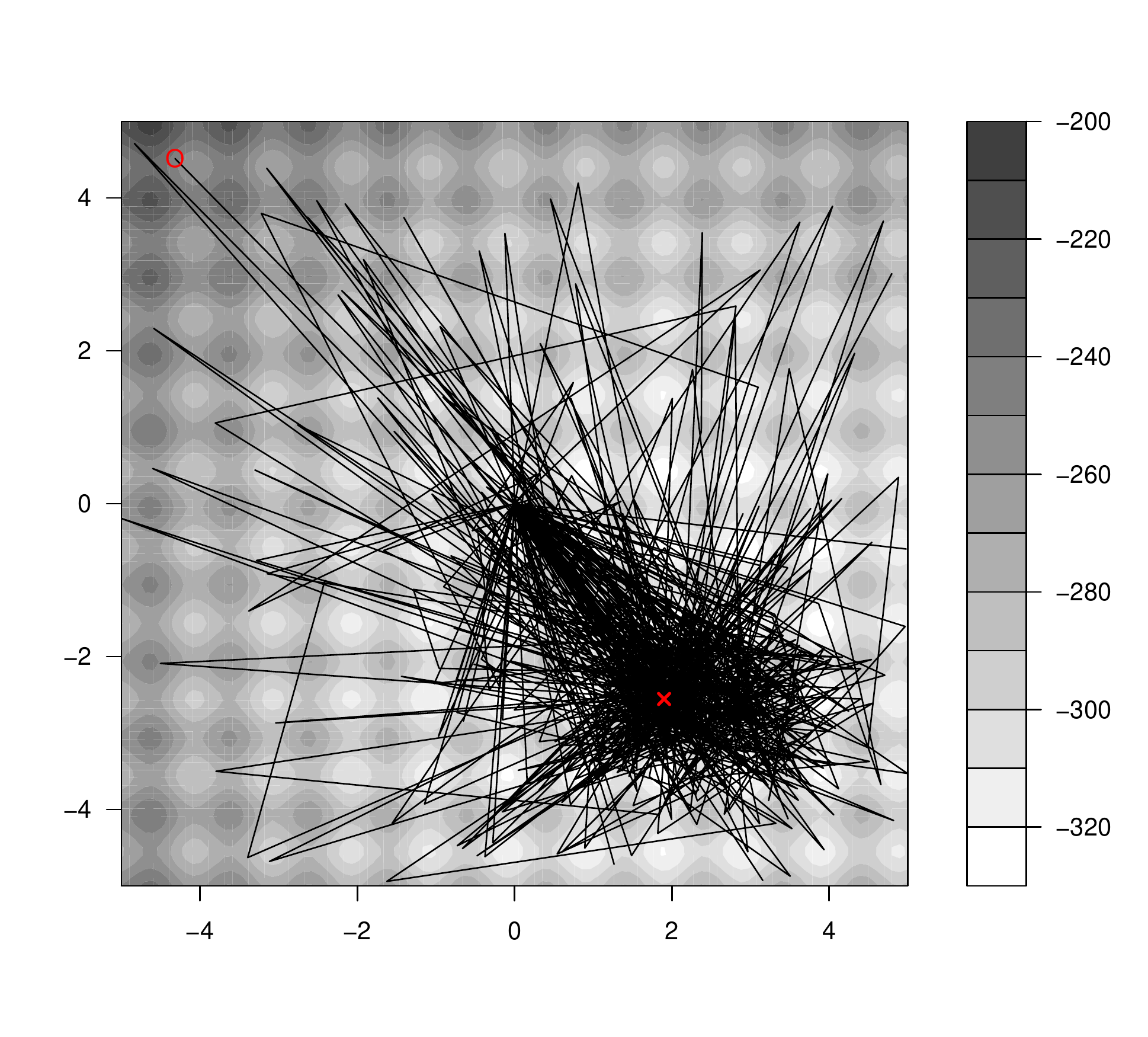}
		\label{fig:trajectories:f1269:f9}
	}\\
\subfloat[Optimising $F_1$, trained on $F_{1,2,6,9}$ (\href{https://youtu.be/nK7NxwkQszg}{movie})] {
	\hspace{-4mm}
	\includegraphics[trim={0 1cm 0 2cm}, width=225pt]{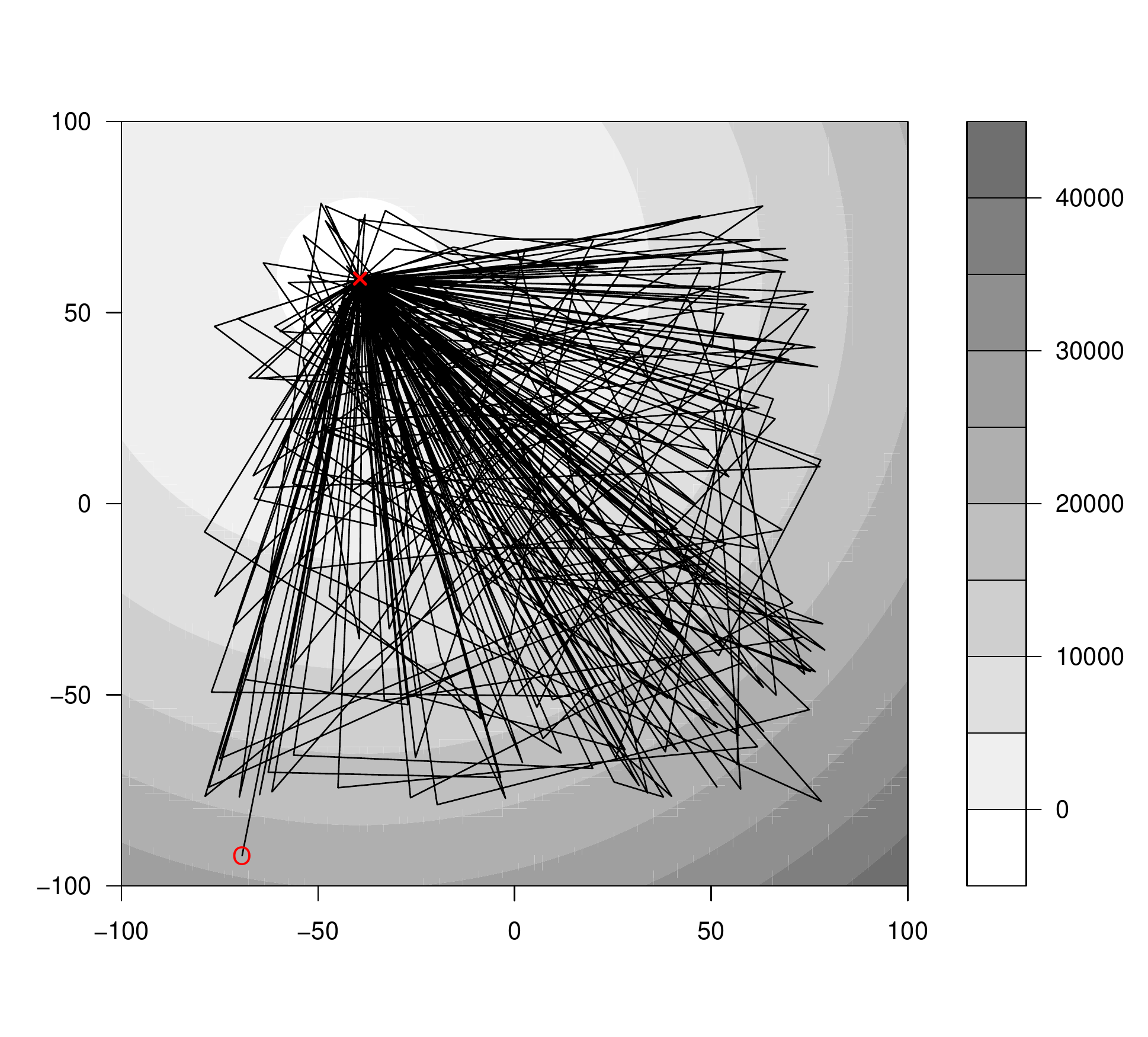}
	\label{fig:trajectories:f1269b:f1}
}
\subfloat[Optimising $F_1$, trained on $F_{1,2,6,9}$ (\href{https://youtu.be/jVlRIEg7vIM}{movie})] {
	\includegraphics[trim={0 1cm 0 2cm}, width=225pt]{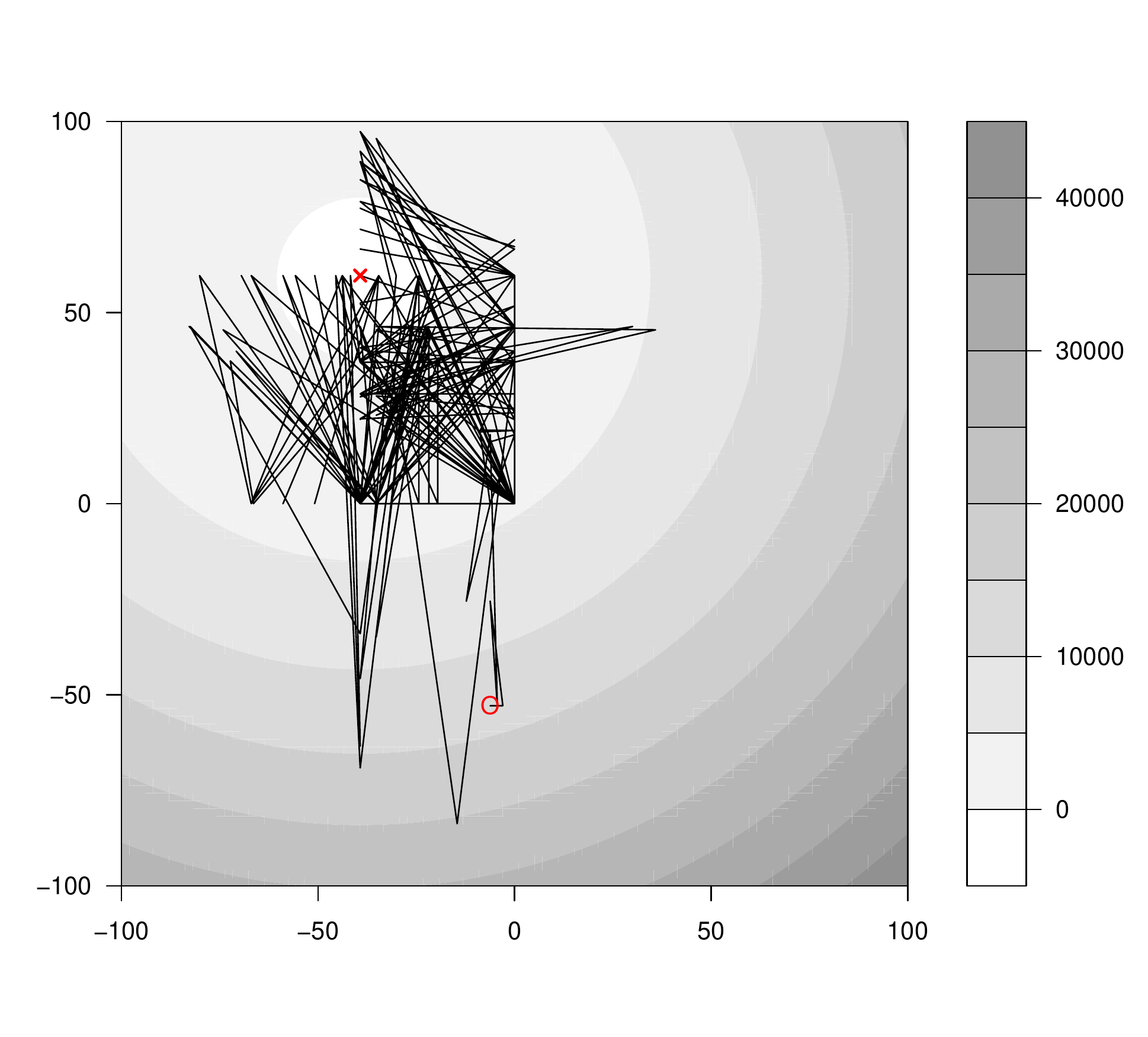}
	\label{fig:trajectories:f1269b:f1b}
}
	\caption{Examples of search trajectories explored by optimisers trained on all four benchmark functions.}
	\label{fig:trajectories}
\end{figure*}

Optimisers are likely to over-learn aspects of the landscapes they are trained on, and this will limit their generality. This can also be seen in Table \ref{table:dimensionality}, which shows how well the evolved optimisers generalise to instances of the same problem with larger (and smaller) dimensionalities; whilst the $F_1$ optimiser generalises well, the other optimisers struggle when the dimensionality is increased. One way of addressing this problem is to train the optimisers on more diverse problems, for example multiple function landscapes or multiple dimensionalities at once. To give an idea of how well this approach works in practice, another set of Push GP runs were carried out with each optimiser evaluated on all four benchmark functions at once, i.e. an optimiser's fitness is the mean of 10 repeats of $F_1$, $F_2$, $F_6$ and $F_9$ from random initial points. Lexicase selection \citep{helmuth2015solving} was used, with each function treated as a separate fitness case; this is to compensate for the large differences in objective values and error rates for each of the benchmark functions, which would otherwise cause the overall mean to be dominated by $F_6$. Since this is essentially a multi-objective problem, there are many equally good solutions to the problem. The error rates for one of these is shown in Table \ref{table:generality}, showing that it does generally well for all functions, but less well than the specialist optimisers for each function. The Push expression is shown in Table \ref{tab:optimisers}. It appears more complex than the others because it uses a loop to rearrange the execution stack. The consequence of this is that sometimes the tangent function is applied to the objective value and a random vector created and added to the current best, as in the $F_1$ optimiser. At other times, the current best is replaced by a random point within the search bounds, i.e. a restart, in optimisation terms. Also, early on in execution, the origin is explored. When applied to $F_1$, this behaviour causes a more irregular trajectory than the optimiser trained only on $F_1$, as shown in Figure \ref{fig:trajectories:f1269:f1}, but it is still effective in reaching the optimum. For $F_9$ (see Figure \ref{fig:trajectories:f1269:f9}), the output of the tangent function causes relatively larger moves due to the smaller bounds of the search space; these are sufficient to hop over local optima basins, but also sometimes small enough to allow search within a basin, and hence it also does fairly well on this function.

A more thorough analysis of these more general optimisers is still to be performed, but an initial inspection of the solutions suggests that they are behaviourally quite diverse. Some of them (such as the one just discussed) behave similarly to the single-function optimisers described earlier. Others behave quite differently. Some examples are shown in Figures \ref{fig:trajectories:f1269b:f1} and \ref{fig:trajectories:f1269b:f1b}, both of which use geometric patterns of movement in order to sample solutions, leading to unusual (yet still effective) search trajectories through the $F_1$ landscape. This suggests that there is still be a lot we could potentially learn from evolved optimisers.

\section{Conclusions} \label{sec:conclusions}

This paper has shown that Push GP can be used to design local optimisers, and do so largely from scratch using primitive instructions. The resulting optimisers tend to be well fitted to the landscapes they were trained on, meaning that they can often approach the optima more rapidly in comparison to a general optimiser that is not aware of the search space characteristics. However, more generality can be achieved by training on multiple landscapes. Although evolved programs are rarely transparent to human understanding, it is often possible to reduce them to compact expressions by removing instructions that have no effect on the output, and then to gain more insight into their effect by observing their behaviour during search. In this respect, they retain an inherent advantage over recent deep learning approaches to designing optimisers, where the translation of trained behaviours into human understanding is much more difficult. Several examples of interpreted evolved Push programs were given in this paper, and showed that they used interesting, and sometimes quite unusual, strategies to explore optimisation landscapes. This supports the idea that this approach may be used to explore optimisation behaviours in a more systematic and problem-dependent manner than traditional ways of designing new optimisation algorithms.

Nevertheless, this is only an initial exploration of the idea. In the experiments, parameters such as the constitution of the function set, the maximum number of instruction executions per move, the maximum number of moves during fitness evaluation, and the population size were fixed. In practice, many of these parameters will have fundamental effects on of how the program space is explored, and further experiments are required to understand these effects. Work is also under way to extend this approach to population-based optimisers.

\bibliographystyle{abbrvnat}
\bibliography{lones_gecco2019} 

\end{document}